\documentclass{article}

\usepackage{PRIMEarxiv}
\usepackage{amsmath}

\usepackage[utf8]{inputenc} 
\usepackage[T1]{fontenc}    
\usepackage{hyperref}       
\usepackage{url}            
\usepackage{booktabs}       
\usepackage{amsfonts}       
\usepackage{nicefrac}       
\usepackage{microtype}      
\usepackage{lipsum}
\usepackage{fancyhdr}       
\usepackage{graphicx}       
\graphicspath{{media/}}     
\usepackage{amsmath} 

\title{DisEmbed: Transforming Disease Understanding through Embeddings
}

\author{
  Salman Faroz \\ 
  \texttt{stsfaroz@gmail.com} \\
}

\begin{document}
\maketitle

\begin{abstract}

The medical domain is vast and diverse, with many existing embedding models focused on general healthcare applications. However, these models often struggle to capture a deep understanding of diseases due to their broad generalization across the entire medical field. To address this gap, I present DisEmbed, a disease-focused embedding model. DisEmbed is trained on a synthetic dataset specifically curated to include disease descriptions, symptoms, and disease-related Q\&A pairs, making it uniquely suited for disease-related tasks. For evaluation, I benchmarked DisEmbed against existing medical models using disease-specific datasets and the triplet evaluation method. My results demonstrate that DisEmbed outperforms other models, particularly in identifying disease-related contexts and distinguishing between similar diseases. This makes DisEmbed highly valuable for disease-specific use cases, including retrieval-augmented generation (RAG) tasks, where its performance is particularly robust.

\end{abstract}
\section{Introduction}
When it comes to understanding diseases, many existing models, such as ClinicalBERT and BioBERT, struggle due to their broad generalization across the medical domain. While these models perform well in general healthcare contexts, they often fail to capture the nuanced relationships between specific diseases and their symptoms. For example, in use cases like Clinical Decision Support, disease diagnosis systems, and disease categorization based on symptoms, these models fall short. They can identify that a given text is related to the medical field, but they often do not understand whether the entities in the text are directly related. 

For instance, while both "brain surgery" and "parkinson's disease" are medical terms, a medical/general model might mistakenly associate them because it treats both as medical concepts, leading to high cosine similarity, even though they are unrelated. To address this gap, I have curated a synthetic dataset focused solely on diseases, where the descriptions and symptoms are not explicitly labeled with symptom names. This forces the model to learn deeper and more precise associations and not rely solely on superficial medical terminology. Although there is an inherent understanding of the correlations between symptoms and diseases, this approach promotes a more focused and accurate understanding of the disease. With DisEmbed, I can cluster small disease sets for retrieval-based use cases, making it particularly valuable for disease-specific tasks. This model represents a step forward in disease embedding, offering a more targeted solution for disease-related applications, where its performance is particularly effective.

\section{Related Work}

Recent research has explored the use of embedding models for semantic search tasks in the medical domain. However, there exists a debate regarding the effectiveness of specialized clinical models compared to general-purpose models. While some studies advocate for domain-specific models, my work challenges this notion.

A recent paper by \cite{excoffier2024generalist} Excoffier and Roehr [2024] investigated the performance of generalist and specialized embedding models for short-context clinical semantic search. Their evaluation methodology involved using ICD-10 codes for evaluation. This approach has limitations. \cite{icd10}ICD-10 codes are primarily used for medical coding and billing purposes, and not designed for natural language processing tasks. They represent diagnoses as discrete codes, lacking the semantic richness required for effective embedding models. The focus on ICD-10 codes potentially biases the evaluation towards models that excel at code matching rather than capturing the nuances of clinical language.

Several existing models cater to the medical domain. NeuML's PubmedBERT base embeddings \cite{neuML}[NeuML] offer a pre-trained model fine-tuned on PubMed literature. MedCodER \cite{li2023medcoder} [Li et al., 2023] presents a generative AI assistant for medical coding tasks.\cite{abhinand202xmedembed} MedEmbed offers a range of pre-trained medical embedding models with varying parameter sizes. These models provide a foundation for medical text analysis, but often lack a specific focus on disease understanding.

The field of medical embeddings has seen significant advancements with the development of models tailored for various healthcare applications. One notable model is SciBERT \cite{beltagy2019scibert}, which is designed for scientific text and has been adapted for biomedical tasks. SciBERT's architecture is based on BERT but trained on a large corpus of scientific literature, making it effective for tasks like named entity recognition and relation extraction in the biomedical domain.Another important contribution is BlueBERT \cite{peng2019transfer}, which combines PubMed abstracts and MIMIC-III clinical notes for pre-training. BlueBERT has shown improved performance in clinical natural language processing tasks, such as clinical concept extraction and medical question answering, by leveraging both biomedical literature and clinical data.The Clinical Transformer \cite{huang2019clinicalbert} is another model that extends BERT for clinical text, focusing on electronic health records (EHRs). It has been fine-tuned for tasks like clinical note classification and patient outcome prediction, demonstrating the potential of transformer-based models in clinical settings.

Additionally, the work by Alsentzer et al. \cite{alsentzer2019publicly} introduced a series of BERT-based models pre-trained on clinical notes, which have been fine-tuned for various clinical NLP tasks. These models highlight the importance of domain-specific pre-training in capturing the unique characteristics of clinical language.

These models collectively underscore the trend towards developing specialized embeddings that cater to the unique requirements of the medical domain, emphasizing the need for models like DisEmbed that focus on specific aspects such as disease understanding.

\section{Methodology}

Before getting into the specifics of dataset creation, it is essential to outline the overarching approach and considerations that guided the development of DisEmbed. The primary objective was to create a model that excels in understanding disease-specific contexts, which necessitated a departure from traditional medical embeddings that often prioritize generalization across the entire medical domain. This focus on diseases required a careful selection of training data and a tailored training strategy to ensure that the model could capture the intricate relationships between diseases and their symptoms. By concentrating on disease-related tasks, DisEmbed aims to fill the gap left by broader models, providing a more nuanced understanding that is crucial for applications like disease diagnosis and symptom mapping. This approach not only enhances the model's performance in disease-specific scenarios but also sets the stage for future advancements in specialized medical embeddings.

\subsection{Dataset Creation}  
For dataset creation, I utilized the International Classification of Diseases (ICD-10-cm) dataset, which contains over 70,000+ disease names. Then these disease names were passed through the GPT-4o-mini model to generate the corresponding symptoms or descriptions, deliberately excluding the disease name in the output. This approach was intended to encourage the model to better understand the underlying disease concepts without relying on the disease names themselves. In addition, also generated question-answer (QA) pairs based on these diseases. During the process, I performed extensive data shuffling and manually removed irrelevant or unrelated entries to improve the quality of the data set. However, given that this data set is synthetic, it may still contain potential biases and caution should be exercised when using it.

To further enhance the model's ability to understand diseases, I ensured that the generated descriptions varied in complexity, ranging from basic symptom lists to more detailed clinical narratives. Although extensive cleaning was performed, some inherent noise in the synthetic data may remain, which could affect the model’s performance in edge cases. The dataset aims to provide a more focused, disease-specific context for model training and can be adapted for various downstream disease-related tasks, such as classification or retrieval-based tasks. This focused approach helps ensure that the model can better capture the nuances of disease understanding while minimizing the impact of unrelated medical terms.
\begin{figure}
    \centering
    \includegraphics[width=0.96\linewidth]{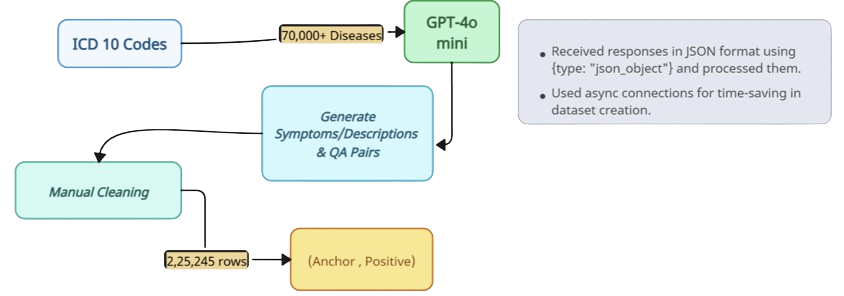}
    \caption{Workflow for Synthetic Dataset Creation Using ICD-10 Codes and GPT-4o-mini}
    \label{fig:loss}
\end{figure}

\subsection{Training Process and Configuration}
For training the DisEmbed model, I utilized that dataset consisting of pairs of anchor and positive examples, where each anchor (a disease description or symptom) is paired with a relevant positive example (a matching disease-related description or symptom). The training objective was to minimize the similarity distance between these paired examples, allowing the model to learn meaningful representations of diseases and their associated symptoms. The loss function used for training was the Multiple Negatives Ranking Loss (MNRL), which is effective in ranking tasks where multiple negative samples may exist for a given positive pair, encouraging the model to focus on distinguishing the correct disease-context relationships.

The model was trained for 4 epochs using the base model BAAI/bge-small-en-v1.5, a lightweight model suited for the task of disease understanding. Training was conducted on an NVIDIA A40 GPU, which provided sufficient computational power for efficient training. Throughout the training process, the model was optimized to ensure it learned strong disease-specific embeddings, which could then be leveraged for downstream tasks such as disease classification, symptom mapping, and retrieval-augmented generation (RAG). Although the model was not trained with a triplet approach, the use of anchor-positive pairs allowed for effective learning of disease-related semantic spaces, yielding promising results in disease-specific tasks.

\section{Model Evaluation}
Evaluating DisEmbed presented a unique challenge due to the lack of dedicated benchmark datasets specifically tailored for disease-focused embedding models. While general medical benchmark datasets are widely available, they primarily cater to broader medical or clinical contexts, which are not the target of DisEmbed. Testing DisEmbed on such datasets would not provide meaningful insights, as the model is specifically designed to excel in disease-related tasks rather than general medical applications.

To address this, I identified disease-specific datasets for evaluation, as they align better with the model's core purpose and training methodology. These datasets, which include disease symptom associations, and Disease pairs, were curated to reflect the nuanced relationships that DisEmbed is designed to capture. The evaluation process focused on tasks such as clustering diseases, identifying disease-related contexts, and distinguishing between similar diseases. These tasks directly assess the model's ability to create meaningful representations of diseases, aligning with its intended use cases.

\subsection{Evaluation Metrics}

To evaluate the performance of DisEmbed, I employed a triplet-based evaluation strategy, using both the `TripletEvaluator` from the `sentence-transformers` library and a custom triplet evaluation implementation for cases where certain pre-trained models, such as ClinicalBERT and BioBERT, were not supported by `sentence-transformers`. The evaluation process is grounded in the triplet learning paradigm, which ensures that the similarity between an anchor and a positive example is greater than the similarity between the anchor and a negative example, within a specified margin.

\subsubsection{Triplet Evaluation Method}
The triplet \cite{henderson2017efficient} evaluation process involves a dataset of triplets $(A, P, N)$, where:
\begin{itemize}
    \item $A$ is the \textit{anchor}, typically a disease description or symptom.
    \item $P$ is the \textit{positive example}, a contextually related entity (e.g., symptoms linked to $A$).
    \item $N$ is the \textit{negative example}, an unrelated entity (e.g., symptoms from a different disease).
\end{itemize}

The similarity function $S(x, y)$ (e.g., cosine similarity) is used to compute the relationship between two embeddings. The evaluation objective ensures that:
\[
S(A, P) > S(A, N) + \text{margin},
\]
where $\text{margin}$ is a tunable parameter to separate positive and negative examples. The accuracy is defined as the percentage of triplets satisfying the above inequality:
\[
\text{Accuracy} = \frac{\text{Number of Correct Triplets}}{\text{Total Number of Triplets}} \times 100.
\]

\paragraph{Implementation using Sentence Transformers.}
The `TripletEvaluator` from the `sentence-transformers` library simplifies this evaluation by leveraging pre-trained models and embeddings. It supports various similarity metrics (e.g., cosine, dot product) and automates the computation of the accuracy for the triplet condition:
\[
S(A, P) > S(A, N) + \text{margin}.
\]
This evaluator integrates seamlessly with models that support tokenization and sentence embeddings, making it a robust choice for standard evaluation scenarios.

\paragraph{Custom Triplet Evaluator for Unsupported Models.}
For pre-trained models like ClinicalBERT and BioBERT, which are not directly supported by the `sentence-transformers` library, the `TripletEvaluator` cannot be used. To address this limitation, a custom triplet evaluator was implemented. This evaluator uses a triplet dataset, a tokenizer, and the model to generate embeddings for $A$, $P$, and $N$. The cosine similarity is then computed between embeddings, and the same triplet condition is evaluated. For this custom implementation:
\begin{itemize}
    \item The embeddings for $A$, $P$, and $N$ are computed as:
    \[
    E(x) = \frac{1}{n} \sum_{i=1}^n x_i,
    \]
    where $x_i$ is the $i$-th token embedding, and $n$ is the number of tokens.
    \item Cosine similarity is calculated as:
    \[
    S(x, y) = \frac{\langle E(x), E(y) \rangle}{\|E(x)\| \cdot \|E(y)\|}.
    \]
    \item Accuracy is computed similarly to the standard triplet evaluator.
\end{itemize}

The custom implementation ensures compatibility with ClinicalBERT, BioBERT, and other models by directly interfacing with tokenized input and performing the embedding and evaluation process.

\subsubsection{Comparison of Approaches}
Both the `TripletEvaluator` and the custom triplet evaluator implement the same core concept of triplet learning. The key differences are in their integration:
\begin{itemize}
    \item The `TripletEvaluator` is optimized for pre-trained models in the `sentence-transformers` framework and offers prebuilt functionality for various similarity metrics and margin configurations.
    \item The custom triplet evaluator is designed for flexibility, allowing the use of models or tokenizers that are not compatible with `sentence-transformers`, such as ClinicalBERT and BioBERT.
\end{itemize}

In practice, both approaches yielded consistent results for models and datasets that supported standard embeddings, confirming the robustness of the triplet evaluation methodology for measuring the semantic understanding of diseases.

\subsection{Benchmarks and Analysis}

To evaluate the performance of DisEmbed-v1, I compared it against a range of existing models on three key datasets: \cite{disease_database} Disease Database,\cite{iclr_medical_diagnosis_dialogue} Medical Diagnosis Dialogue (MDD), and \cite{cod_patient_symdisease} CoD-PatientSymDisease. Table~\ref{tab:model_performance} summarizes the results. 

\begin{table}[htbp]
    \centering
    \renewcommand{\arraystretch}{1.34} 
    \large 
    \resizebox{\textwidth}{!}{ 
    \begin{tabular}{lcccc}
        \hline\hline
        \textbf{Model} & \textbf{Parameters} & \textbf{Disease Database (\%)} & \textbf{Medical Diagnosis Dialogue (\%)} & \textbf{CoD-PatientSymDisease (\%)} \\
        \hline
        PubMedBERT - NeuML     & 110M & 88.2 & 78.6 & 85.3 \\ 
        ClinicalBERT           & 110M & 69.6 & 59.7 & 65.5 \\ 
        MedEmbed-Base-v0.1     & 110M & 92.8 & 87.1 & 92.6 \\ 
        MedEmbed-Small-v0.1    & 33M  & 89.1 & 89.9 & 87.9 \\ 
        MedEmbed-Large-v0.1    & 325M & 92.8 & 87.1 & 92.6 \\ 
        Bio-ClinicalBERT       & 110M & 71.6 & 60.2 & 65.6 \\ 
        biobert-v1.1           & 110M & 71.3 & 62.6 & 65.7 \\     
        Snowflake-Arctic-M     & 110M & 85.4 & 89.5 & 81.5 \\ 
        BAAI-bge-large-en-v1.5 & 335M & 91.5 & 84.9 & 90.3 \\ 
        \textbf{DisEmbed-v1 }   & 33M & \textbf{94.5} & \textbf{91.6} & \textbf{93.7} \\ 
        \hline\hline
    \end{tabular}
    }
    \caption{Performance comparison of various models across datasets.}
    \label{tab:model_performance}
\end{table}

\subsubsection{Dataset Descriptions}
\begin{itemize}
    \item \textbf{Disease Database:} This dataset consists of 9.62k rows and focuses on mapping symptoms to their associated diseases. It serves as a strong benchmark for evaluating models in tasks related to disease and symptom prediction.
    \item \textbf{CoD-PatientSymDisease:} This dataset includes both explicit and implicit symptoms, along with the target disease (39.1k rows). It tests the model's ability to understand nuanced relationships between symptoms and diagnoses.
    \item \textbf{Medical Diagnosis Dialogue (MDD):} Introduced in the ICLR 2021 Workshop on Machine Learning for Preventing and Combating Pandemics, this dataset includes 12 diseases in the general medical domain. It is based on the \cite{muzhi_dataset} MuZhi dataset, where source medical records were converted into structured user goals consisting of disease tags, explicit symptoms, and implicit symptoms, ensuring privacy preservation. The MuZhi dataset itself provides sequences of symptoms and corresponding diagnoses, making it suitable for training models on automatic diagnosis via symptom sequence generation.
\end{itemize}

\subsubsection{Performance Analysis}
The results in Table~\ref{tab:model_performance} highlight the better performance of DisEmbed-v1 across all datasets. With a smaller parameter size (33M), DisEmbed-v1 achieves state-of-the-art results, outperforming larger medical-specific models like PubMedBERT, \cite{clinicalbert}ClinicalBERT, and \cite{bio-clinicalbert}Bio-ClinicalBERT, as well as generic models like \cite{snowflake_arctic}Snowflake and \cite{bge}BAAI-bge-large-en-v1.5. This demonstrates that DisEmbed-v1 is highly effective for disease-related tasks, particularly in scenarios requiring accurate mapping of symptoms to diseases.

\subsubsection{Limitations}
While the benchmarks confirm DisEmbed-v1's exceptional performance on disease-related tasks, its scope is limited to medical contexts due to its disease-focused training. Although the comparison includes two generic models, Snowflake-Arctic-M and BAAI-bge-large-en-v1.5, the results do not generalize to broader medical benchmarks. Other medical models trained on a wider variety of tasks may outperform DisEmbed-v1 in domains beyond disease-specific use cases.

\section{Experimental Results and Analysis}

In this section, I present a detailed analysis of the DisEmbed model's performance in comparison to other state-of-the-art models, such as PubMedBERT, MedEmbed and BioBERT. The focus of this analysis is on the model's ability to accurately capture disease-specific contexts, as demonstrated through cosine similarity evaluations.

\subsection{Cosine Similarity Evaluation}

To assess the semantic understanding of diseases by DisEmbed, I conducted a series of experiments using cosine similarity as a metric. The experiments involved comparing the embeddings of a symptom description with those of two diseases: \textit{neuropathy} and \textit{epilepsy syndrome}. The symptom description used was:

\begin{quote}
\textit{"Reduced sensation in the hands and feet, along with tingling or numbness, is often accompanied by decreased reflexes and muscle weakness."}
\end{quote}

This description is indicative of neuropathy, a condition characterized by nerve damage, but not of epilepsy syndrome, which involves seizures and neurological symptoms unrelated to peripheral nerve damage.

\subsubsection{Model Comparisons}

The cosine similarity results for DisEmbed and other models are summarized in Table~\ref{tab:cosine_similarity}. These results highlight the model's ability to distinguish between related and unrelated disease contexts.

\begin{table}[htbp]
    \centering
    \begin{tabular}{lcc}
        \hline\hline
        \textbf{Model} & \textbf{Cosine Similarity with Neuropathy} & \textbf{Cosine Similarity with Epilepsy Syndrome} \\
        \hline
        \textbf{DisEmbed-v1} & 0.6333 & 0.1062 \\
        MedEmbed-base-v0.1 & 0.7024 & 0.5724 \\
        BAAI-bge-large-en-v1.5 & 0.6719 & 0.5203 \\
        PubMedBERT & 0.5864 & 0.1845 \\
        BioBERT & 0.8309 & 0.8300 \\
        \hline\hline
    \end{tabular}
    \caption{Cosine similarity results for symptom description with disease embeddings.}
    \label{tab:cosine_similarity}
\end{table}

\subsubsection{Analysis of Results}

The results demonstrate that DisEmbed-v1 exhibits a significantly lower cosine similarity for the unrelated disease (\textit{epilepsy syndrome}) compared to the related disease (\textit{neuropathy}). This indicates a strong ability to differentiate between diseases based on symptom descriptions, which is crucial for accurate disease identification and retrieval tasks.

In contrast, models like \cite{biobert}BioBERT show high cosine similarity for both diseases, suggesting a less nuanced understanding of disease-specific contexts. This behavior can lead to incorrect associations in retrieval tasks, where distinguishing between similar but distinct diseases is essential.

\subsection{Implications for Disease Retrieval}

The ability of DisEmbed to maintain low cosine similarity for unrelated diseases suggests its potential effectiveness in retrieval-augmented generation (RAG) tasks. By ensuring that embeddings for unrelated diseases are not closely aligned, DisEmbed can enhance the accuracy of retrieval systems that rely on embedding databases. This is particularly valuable in clinical decision support systems, where precise disease identification is critical.

Furthermore, the experiment highlights the importance of constructing a well-curated embedding database. With accurate disease names and descriptions, DisEmbed can significantly improve retrieval processes, offering a robust solution for disease-specific applications.

\section{Future Work and Improvements}

\subsection{Enhancing Dataset Quality and Model Parameters}
While DisEmbed has demonstrated strong performance in disease-specific tasks, there is potential for further improvement by enhancing the dataset's quality and size. Future work will focus on expanding the dataset to include a broader range of diseases and symptoms, ensuring a more comprehensive representation of the medical domain. Additionally, increasing the model's parameters could further improve its ability to capture complex disease relationships, while maintaining efficiency in disease-specific contexts.

\subsection{Balancing Disease-Specific and General Medical Domains}
Although DisEmbed excels in disease-specific applications, it is crucial to ensure that its performance in general medical domains is not compromised. Future iterations of the model will aim to balance disease-specific understanding with broader medical knowledge, enabling it to perform well across a variety of medical tasks. This will involve fine-tuning the model on diverse medical datasets to enhance its versatility.

\section{Model and Dataset Availability}

The DisEmbed model, trained on the curated dataset, is publicly available for research and development purposes. Researchers and practitioners can access the model at the following link: \url{https://huggingface.co/SalmanFaroz/DisEmbed-v1}. 

The dataset used for training, which includes symptom-disease pairs, can be accessed at \url{https://huggingface.co/datasets/SalmanFaroz/DisEmbed-Symptom-Disease-v1}. 

These resources provide a foundation for further exploration and application in disease-related tasks.

\section{Conclusion}

The experimental results demonstrate the efficacy of DisEmbed in capturing disease-specific nuances, even with its compact model size. Its better performance in distinguishing related and unrelated diseases highlights its potential as a tool for medical applications requiring precise disease characterization. Future work will focus on expanding the embedding database and incorporating additional disease contexts to further enhance its utility. Moreover, efforts will be directed toward optimizing its computational efficiency to ensure seamless integration into real-world clinical workflows.

\bibliographystyle{unsrt}  
\bibliography{references}  

\end{document}